\documentclass{article}
\usepackage{spconf,amsmath,graphicx}

\usepackage{enumitem}
\setlist{nosep, leftmargin=14pt}

\usepackage{mwe} 
\usepackage{amssymb}
\usepackage{comment}
\usepackage{amsmath}
\usepackage{arydshln}
\usepackage{booktabs}
\usepackage{bm}

\title{Cluster Entropy: Active Domain Adaptation\\ in Pathological Image Segmentation
}

\name{
Xiaoqing Liu$^{1}$,
Kengo Araki$^{1}$,
Shota Harada$^{1}$,
Akihiko Yoshizawa$^{2}$,
Kazuhiro Terada$^{2}$,
Mariyo Kurata$^{2}$,}
\secondlinename{
Naoki Nakajima$^{2}$,
Hiroyuki Abe$^{3}$,
Tetsuo Ushiku$^{3}$,
Ryoma Bise$^{1}$
}
\address{
$^{1}$ Kyushu University, Fukuoka, Japan, 
$^{2}$ Kyoto University, Kyoto, Japan, \\
$^{3}$ The University of Tokyo, Tokyo, Japan
}

\begin{document}

\maketitle
\begin{abstract}
The domain shift in pathological segmentation is an important problem, where a network trained by a source domain (collected at a specific hospital) does not work well in the target domain (from different hospitals) due to the different image features.
Due to the problems of class imbalance and different class prior of pathology, typical unsupervised domain adaptation methods do not work well by aligning the distribution of source domain and target domain.
In this paper, we propose a cluster entropy for selecting an effective whole slide image (WSI) that is used for semi-supervised domain adaptation. This approach can measure how the image features of the WSI cover the entire distribution of the target domain by calculating the entropy of each cluster and can significantly improve the performance of domain adaptation.
Our approach achieved competitive results against the prior arts on datasets collected from two hospitals.
\end{abstract}
\section{Introduction}
\label{sec:intro}

In pathological segmentation, which segments the regions of cancer subtypes in
a whole slide image (WSI), the domain shift problem still remains a challenging
problem. 
Even though a large amount of labeled data is collected from a specific
hospital (source domain) and used to train a deep neural network, the network
does not often work well for data collected from a different hospital (target
domain) due to domain shift. 
The staining materials, imaging devices, and tissue-cutting operations often differ among hospitals; thus, the image features are different, such as hue, saturation, and brightness. 
Such differences among image feature distributions are called domain shift problems.

To cope with domain shift problem, many domain adaptation methods have been proposed. 
The purpose of unsupervised domain adaptation is to align the feature distribution of the target data ({\it e.g.}, unsupervised data collected from different hospitals) with that of the source data ({\it i.e.}, original supervised training data).
To align the distributions, adversarial techniques have often been used \cite{ganin2016domain,Tzeng_2017_CVPR,hoffman2018cycada,yan2017mind,kang2019contrastive,saito2018maximum}. 
These methods have a limitation in medical image analysis since domain shift problems in this field are more complex. 
The distribution of tumors may sometimes change with different hospitals, and a larger hospital tends to handle more severe patients than a small one ({\it i.e.}, the class prior and distributions may differ between target and source domains).
In this case, even if entire distributions of domains are aligned, the class distributions in both domains may not be aligned.
\begin{figure}[t]
\centering
\includegraphics[width=8.5cm]{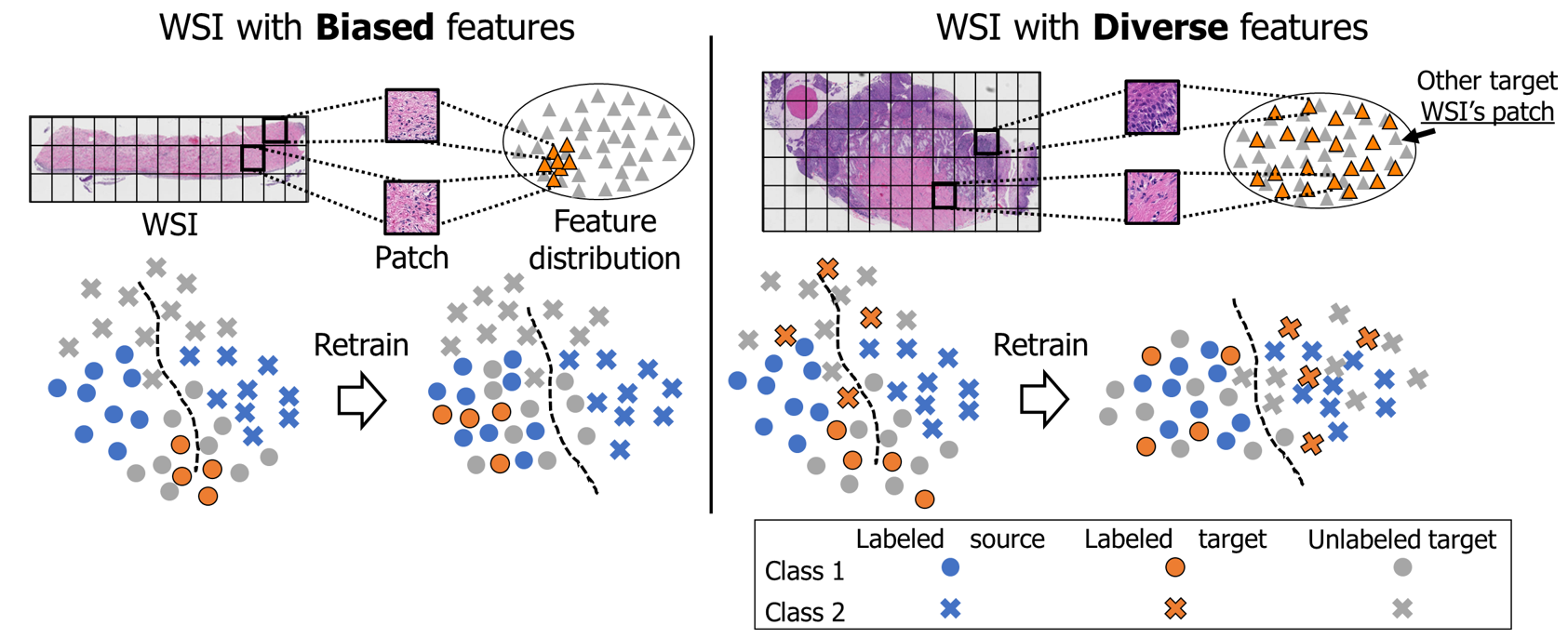}
\caption{Left: example of un-effective WSI, which has biased features. Right: example of effective WSI, which has diverse features and its distribution covers entire distribution of target domain.}
\label{fig:idea}
\end{figure}

Additional annotations for the target domain are helpful in solving such domain shift problems. 
Several semi-supervised domain adaptation methods use a small amount of labeled data in the target domain \cite{Saito_2019_ICCV,kim2020attract}.
These methods assume that additional labeled data in the target domain are initially given and do not consider how the data is selected.

The effectiveness of the semi-supervised domain adaptation often depends on the initial labeled data in the target domain.
As shown in Fig. \ref{fig:idea}, if the labeled data in the feature distribution is biased, where it often occurs in small sampling, retraining is not well performed.
In contrast, the labeled data that covers the entire distribution of the target data makes retraining well. 
It is important to select effective WSIs automatically in semi-supervised domain adaptation to achieve a time-saving and excellent performance.
Therefore, we choose to apply active learning (sample selection method for additional annotation) for semi-supervised domain adaptation. 

To select an effective WSI from the target data for active domain adaptation, we propose a metric called cluster entropy to measure the diversity of patch features of a single WSI in a distribution of all target data.
The WSI with the highest cluster entropy is selected among all WSIs in the target data to be additionally annotated.
Then, we retrain the network using the source data and the selected WSI.
Retraining using data that covers the distribution of the target domain forces the distribution of both the source and the target to be similar, which improves the classification performance.
The most significant difference of our method from typical active learning is that we consider the distribution of a set of patch images instead of a single sample since a WSI is often treated as a set of small patches due to its large size. 
Our approach is verified on real pathological datasets collected from two hospitals and significantly improves the performance.

\noindent
{\bf Related work about active learning:}
Active learning, a method for selecting highly effective data from unlabeled data, has been proposed to improve classification performance with a low annotation cost.
Many methods follow pool-based sampling, which selects data using a model pre-trained on a small number of labeled data \cite{gal2017deep,ducoffe2018adversarial,sinha2019variational,zhdanov2019diverse,prabhu2021active,singh2021improving}.
There are two main approaches for pool-based sampling: uncertainty sampling, which selects samples with high uncertainty using a Bayesian CNN \cite{gal2017deep}, decision boundaries \cite{ducoffe2018adversarial}, and diversity sampling, which selects a wide variety of samples from a dataset \cite{zhdanov2019diverse,singh2021improving}. 
However, these methods do not consider a selection metric for a set of instances (patches) corresponding to a WSI.

\section{Active domain adaptation by cluster entropy}
\label{sec:format}

The purpose of our study is to select an effective WSI from unlabeled WSIs in a target domain to be additionally annotated.
Retraining the network using an effective WSI improves the performance in the target domain.

\subsection{Problem setup}
A set of labeled source images $\mathcal{D}_s=\{\bm{X}_i^s, \bm{S}_i^s\}_{i=1}^{m_s}$ and a set of unlabeled target images $\mathcal{D}_t = \{\bm{X}_i^t\}_{i=1}^{m_t}$ are given, where $\bm{X}_i^s$, $\bm{X}_i^t$ are WSIs in the source and target domains, respectively, $\bm{S}_i^s$ is the corresponding ground truth of the semantic segmentation, where each pixel has a class.
We separate the giga-pixel WSIs into patch images $\mathcal{D}'_s = \{\bm{x}_j^s, y_j^s\}_{j=1}^{N_s}$ for the labeled data, and $\mathcal{D}'_t = \{\bm{x}_j^t\}_{j=1}^{N_t}= \{\bm{X}_i^t=\{\bm{x}_{j}^t\}_{j=1}^{n_i^t}\}_{i=1}^{m_t}$ for the unlabeled data. $\bm{x}_j$ is a cropped patch image from WSIs and $y_j$ is the ground truth of the class. $N_s$, $N_t$ are the total number of patches for the source and target, respectively, and $n_i^t$ is the number of patches in the $i$-th WSI.
In active learning for domain adaptation in pathology, we select a WSI (a set of patches) from $\mathcal{D}'_t$. Pathologists then annotate the boundaries of each class in the selected WSI. The additionally labeled patches in the target domain are described as $\mathcal{A}_t = \{\bm{x}_j, y_j\}_{j=1}^{n_i^t}$.
Then, $\mathcal{A}_t$ is used to re-train the network.
This problem setup is different from the typical active learning methods, which select individual instances in all unlabeled data or select pixels from an entire image.

\subsection{What is a good selection of labeled target for semi-supervised domain adaptation}
Pathological tissue is cut from the human body, and there are individual differences in the shape, hue, saturation, brightness, and cell density of tissue sections.
Therefore, the patch images in the target domain $\mathcal{D}'_t$ have various features.
To extract image features of each patch, we use the CNN $f_c$ that has been trained using the source data $\mathcal{D}'_s$. The set of the extracted features are described as $\mathcal{F}_t=\{f_c(\bm{x}_j^t)\}_{j=1}^{N_t}$ for the target domain, and $\mathcal{F}_s=\{f_c(\bm{x}_j^s)\}_{j=1}^{N_s}$ for the source domain.

When we observe the feature distribution of patches $\{f_c(\bm{x}_j^t)\}_{j=1}^{n_i^t}$ in a single WSI $\bm{X}_i^t$, the variation of patch features is very different among WSIs.
For example, the patches of the left image in Fig. \ref{fig:idea} contain the similar appearance patches, whose distribution is biased from the entire target domain ($\{f_c(\bm{x}_j^t)\}_{j=1}^{n_i^t} \nsim \mathcal{F}_t$). 
Only specific classes may be contained, and some class samples are lacking. 
Such biased patch samples in retraining often affect the performance as shown in Fig. \ref{fig:idea}(Left). In this case, only the samples around the biased labeled samples can be aligned in the feature space by training the source and the labeled target together, but not for the other samples that are far from the labeled ones.

In contrast, the right image in Fig. \ref{fig:idea} contains diverse features, in which the patch features of the WSI cover the entire distribution of all the target domain ($\{f_c(\bm{x}_j^t)\}_{j=1}^{n_i^t} \sim \mathcal{F}_t$). Such diverse patches are likely to contain samples of all classes.
As shown in Fig. \ref{fig:idea}, the representation learning using the data that covers the target domain forces the distribution of both the source and the target to be similar. This improves the classification performance.
Therefore, we propose a method for selecting a WSI that has diverse patch features.

\subsection{Cluster entropy}
We aim to select a WSI whose patch features cover the entire distribution of all the target domain. For this purpose, we propose a selection method based on cluster entropy.
Here, cluster entropy is a measure of whether patches in a single WSI are uniformly distributed in the distribution of the target domain. In other words, a WSI with a large cluster entropy indicates that it has a variety of features.

Fig.\ref{fig:proposed_method} shows the overview of the method for calculating cluster entropy. Clustering is performed on all patch features $\mathcal{F}_t$ of the target data.
Since the extracted feature vectors have a high dimension, we perform Principal Component Analysis (PCA) to reduce the dimension before clustering, where we use k-means++\cite{arthur2006k}. After that, the cluster entropy $H$ for each target WSI is calculated as:

\begin{figure}[t]
\centering
\includegraphics[width=\linewidth]{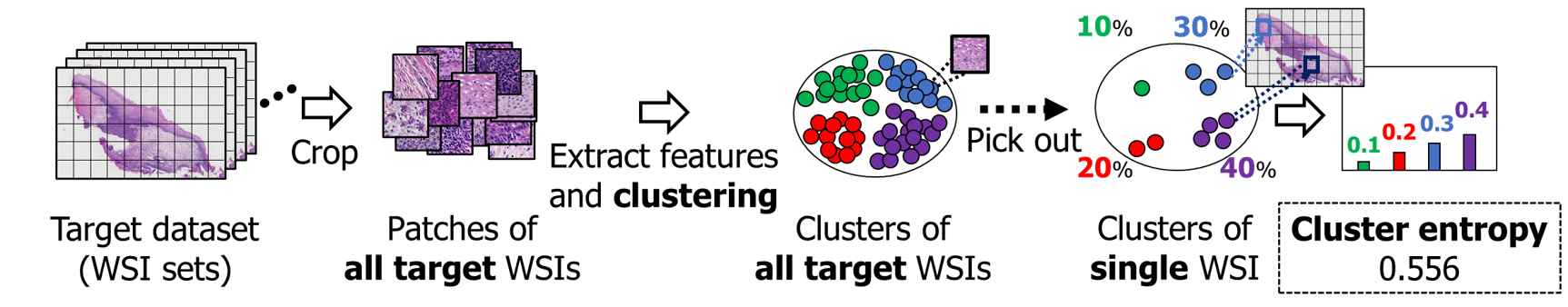}
\caption{Calculation method of cluster entropy. First, feature extraction is performed for all the target data using CNN trained using the source data.  Then, clustering is performed using k-means++\cite{arthur2006k}. After that, the proportion of clusters is calculated for each WSI and the entropy is calculated.
}
\label{fig:proposed_method}
\end{figure}

\begin{equation}
    H = -\sum_{i=1}^{K}P(i)~logP(i),~P(i) = \frac{C_{i}}{\sum_{i=1}^{K}C_{i}},
    \label{eq:cluster_entropy_eq}
\end{equation}
where, $K$ is the total number of clusters, $C_{i}$ is the number of samples in a cluster $i$, and $P(i)$ is the proportion of all samples in a cluster $i$. If the features of the selected WSI are uniformly distributed in all clusters, the cluster entropy takes the maximum value. If the features are in a specific cluster, the cluster entropy tends to take a lower value.  Therefore, the diversity of the features can be measured.

In the proposed method, the WSI with the highest cluster entropy is selected from all the WSIs in the target data. Then, the selected WSI is annotated and the network is retrained with the source WSIs and the newly annotated WSI $\mathcal{A}_t$. In this way, the feature distributions of the source and target domains are made closer together.

\section{Experiments}
To verify the effectiveness of our cluster entropy, we compared the performance after retraining using a large amount of source data and additionally added supervised data in the target domain by changing the additional WSI with different cluster entropy. 
As a baseline, we directly used a network trained by source data without using target data (experiment {\bf S$\rightarrow$T}).

\subsection{Experimental Setting}
\noindent
\textbf{Dataset:}
In this experiment, we used private datasets of WSIs collected from two hospitals (247 WSIs for the source domain, and 108 WSIs for the target domain), where most of the WSIs (233+96) were captured from a different patient.
We classified three classes of cervical cancer stages: non-neoplasm (Non-Neop.), low squamous intraepithelial lesion (LSIL), and high squamous intraepithelial lesion (HSIL).
To show the robustness of the training data in the source domain, we conducted 5-fold cross-validation, where the 247 source WSIs were randomly split into 5 sets in advance. In each validation, 80\% of the WSIs in four of the sets were randomly assigned to the training data and the remaining 20\% to the validation data. The remaining one set was not used in each validation.
The validation data of the source domain was used as a criterion to stop the optimization of experiment S$\rightarrow$T, which was trained only with the source data.

In experiments, we selected 15 WSIs for comparison based on cluster entropy.
The 108 WSIs in the target domain were sorted with the cluster entropy, and the highest 5 WSIs ({\bf High}), which are our method, the middle 5 WSIs ({\bf Med}), and the lowest 5 WSIs ({\bf Low}) were then selected as the labeled data.
Each of them was used as training data with plenty of source WSIs with 5-fold cross-validation. In total, $15\times5$ experiments were performed.
For the validation data, 20 target WSIs were randomly selected from the rest.
The remaining 73 target WSIs were used for the test in every method. After separating the training, validation, and test data, a WSI was cut into 256$\times$256 patch images with 40x magnification in all experiments, in which the source data contained 163,877 patches, and the target data contained 76,443 patches.

There was a class imbalance among the three classes (i.e., Non-Neop.: 146,524, LSIL: 11,707, HSIL: 5,646 patches for source data). To cope with this imbalance, the patches were randomly increased (over-sampling) from the minority class and decreased (under-sampling) from the majority class at each epoch. During the training time, the patches were randomly flipped vertically and horizontally for data augmentation.

\noindent
\textbf{Implementation details:}
ResNet-50 \cite{he2016deep} was used for the CNN $f_c$, and Adam \cite{kingma2014adam} was used for the optimization function with a learning rate of $10^{-3}$. For the initial weights of the CNN, weights trained by ImageNet \cite{deng2009imagenet} were used in S$\rightarrow$T, and the weights pre-trained in S$\rightarrow$T were used in the comparative methods and our method.
Before calculating cluster entropy, we reduced the dimension of the feature vector from 2048 dimensions to 30 dimensions by PCA. In this experiment, we set the number of clusters to $K=10$. 
The batch size was set to 32 in S$\rightarrow$T and 64. In our method, we adjusted the batch size so that each batch contained 32 patches of the source and target to prevent the influence of imbalance between domains.
After early stopping, the weights with the highest mIoU in validation was used for testing.

\noindent
\textbf{Metrics:}
We used the follow metrics: the mean of the per-class precision (mPrecision), mean of the per-class recall (mRecall), mean of the per-class dice-coefficient (mDice), and mean of the intersection over union (mIoU), where mDice and mIoU are the most important metrics.
The average and standard deviation were calculated for each method.

\subsection{Main Results}
\label{sec:results}

\begin{table}[t]
    \centering
    \caption{Evaluation results. Cluster Entropy shows the average of metrics in each condition respectively; Lowest~(Low)~, Medium~(Med)~, and Highest~(High)~. 
    * indicates $p<0.05$ in statistical test.}
     \label{tab:eval_average}
     \resizebox{\columnwidth}{!}{
     \begin{tabular}{l|cccc}
     \hline
     Experiment
     & mPrecision& ~mRecall~  & ~mDice~  & ~mIoU~~ \\ 
     \hline \hline
     S $\rightarrow$ T 
     & 0.525{\scriptsize $\pm$0.028}
     & \textbf{0.686}{\scriptsize $\pm$0.057} 
     & 0.508{\scriptsize $\pm$0.068}
     & 0.406{\scriptsize $\pm$0.058}\\
     DANN \cite{ganin2016domain}
     & 0.482{\scriptsize $\pm$0.028}
     & 0.663{\scriptsize $\pm$0.033}
     & 0.475{\scriptsize $\pm$0.037}
     & 0.366{\scriptsize $\pm$0.037}\\
     Multi-anchor \cite{ning2021multi} 
     & 0.511{\scriptsize $\pm$0.058}
     & 0.517{\scriptsize $\pm$0.071}
     & 0.509{\scriptsize $\pm$0.046}
     & 0.423{\scriptsize $\pm$0.040}\\
     Cluster Entropy (Low)       
     & 0.517{\scriptsize $\pm$0.038}
     & 0.649{\scriptsize $\pm$0.040}  
     & 0.505{\scriptsize $\pm$0.070}
     & 0.408{\scriptsize $\pm$0.064}\\
     Cluster Entropy (Med)   
     & 0.549{\scriptsize $\pm$0.052}
     & 0.611{\scriptsize $\pm$0.080}
     & 0.534{\scriptsize $\pm$0.050}  
     & 0.438{\scriptsize $\pm$0.045}\\
     Cluster Entropy (High)  
     & \textbf{0.563}{\scriptsize $\pm$0.034}  
     & 0.662{\scriptsize $\pm$0.047}
     & *\textbf{0.576}{\scriptsize $\pm$0.025}  
     & *\textbf{0.474}{\scriptsize $\pm$0.024} \\ 
     \hline
     T $\rightarrow$ T   
     & 0.571{\scriptsize $\pm$0.054}          
     & 0.651{\scriptsize $\pm$0.060}       
     & 0.590{\scriptsize $\pm$0.049}          
     & 0.507{\scriptsize $\pm$0.056}          
     \\ 
     \hline
    \end{tabular}
    }
\end{table}

We compared our method ({\bf Cluster Entropy}) in the order of cluster entropy ({\bf High, Med, Low}) with two prior arts. We applied {\bf DANN \cite{ganin2016domain}} which is an existing unsupervised adversarial domain adaptation.
{\bf Multi-anchor} is one of the state-of-the-art methods of active domain adaptation, which selects active samples by calculating L2 distance from target domain to anchors of source domain in \cite{ning2021multi}. We applied it to select 5 active target samples every fold for retraining. We also show the results of {\bf T$\rightarrow$T}, where the target data was estimated by the network trained by the same domain. Here, this task was difficult due to class imbalance.

Table.\ref{tab:eval_average} shows the average of these metrics for each method, where the average was calculated from 25 experiments for each (5 selected target WSIs with 5-fold cross-validation) in  Cluster Entropy.
Our method, Cluster Entropy (High), achieved the highest performances in mDice, mIoU, in which (High) was significant ($p<0.05$) in Tukey's test compared to (Med), (Low) and the comparative methods.
In addition, in the order of cluster entropy (High, Med, Low), the evaluation values of mDice and mIoU were also higher.
This shows that the proposed cluster entropy was correlated with the segmentation performance.
The performances of mDice and mIoU in (High) were close to the oracle ({\bf T$\rightarrow$T}), where the network was trained in the same domain, even though our method used only a single WSI in the target domain.

\noindent
\subsection{Visualization}
\begin{figure}[t]
\centering
\includegraphics[width=8.5cm]{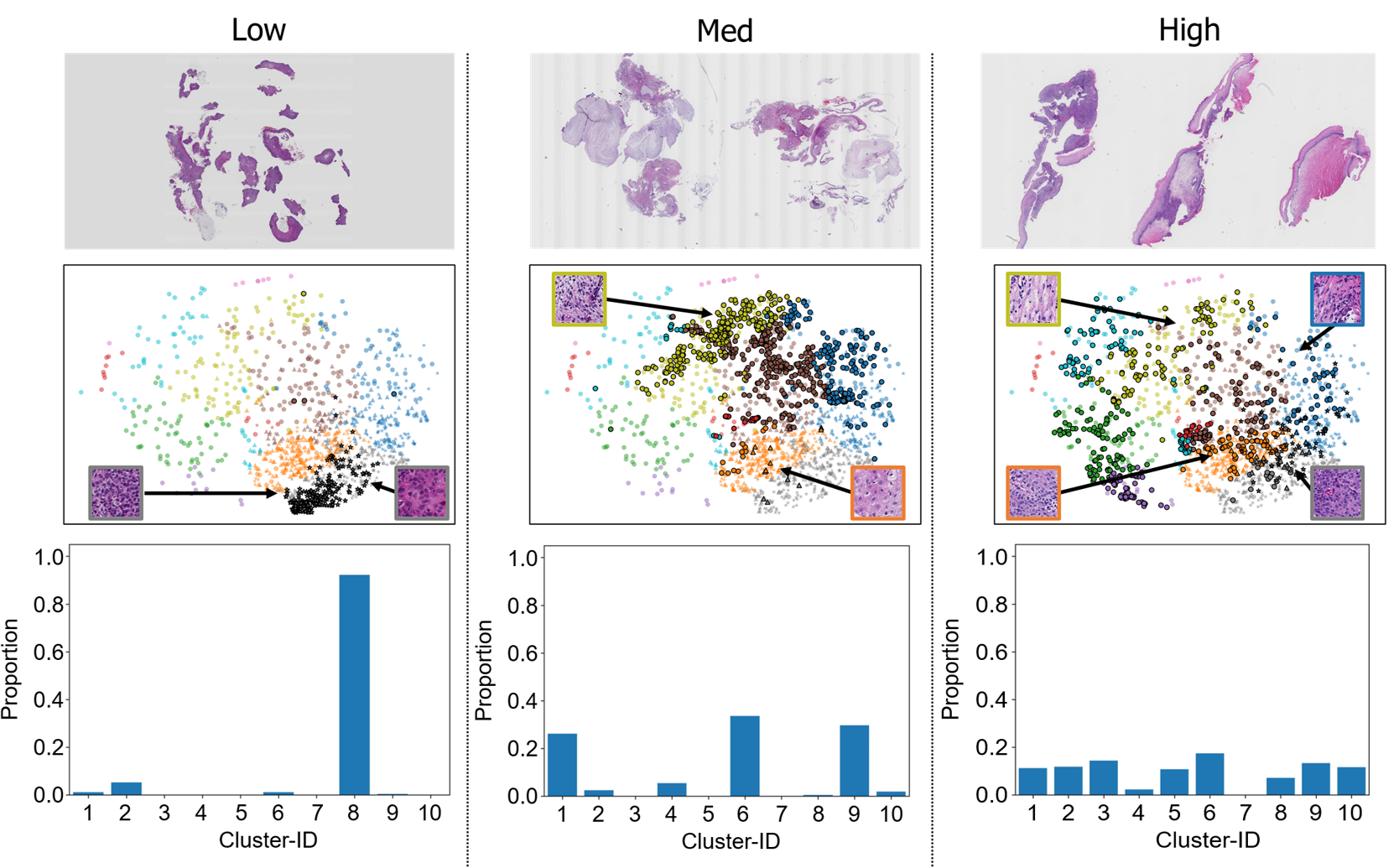}
\caption{Example of WSIs with different cluster entropy. Columns from left to right show WSIs whose cluster entropy was the lowest, medium, and highest, respectively.
Rows from top to bottom represent original image, feature distribution, and histogram of cluster to which patches belong. In feature distribution, color and shape indicate cluster-ID and class. Highlighted markers represent patches of selected WSI.}
\label{fig:labeled_target_WSI}
\end{figure}

Fig. \ref{fig:labeled_target_WSI} shows examples of WSIs whose cluster entropy was the lowest, medium, and highest among the 108 WSIs in the target data. From high to low, the figures represent the original WSI, the feature distribution, and the histogram of the cluster to which the patches cropped from the WSI belong.
We can observe that the distribution of the lowest cluster entropy was very biased and that most patches belonged to cluster 8.
Thus, the WSI had a low cluster entropy.
The WSI of the medium cluster entropy was distributed more widely than the minimum, but some classes contained no or very few patches. 
In contrast, the WSI with the highest cluster entropy contained various features, and these covered the entire distribution of the target domain. From the results, all the clusters contained uniform patches, and the cluster entropy was the highest.
These results show that our metric can measure the diversity of a feature distribution properly.

\begin{figure}[t]
\centering
\includegraphics[width=8.5cm]{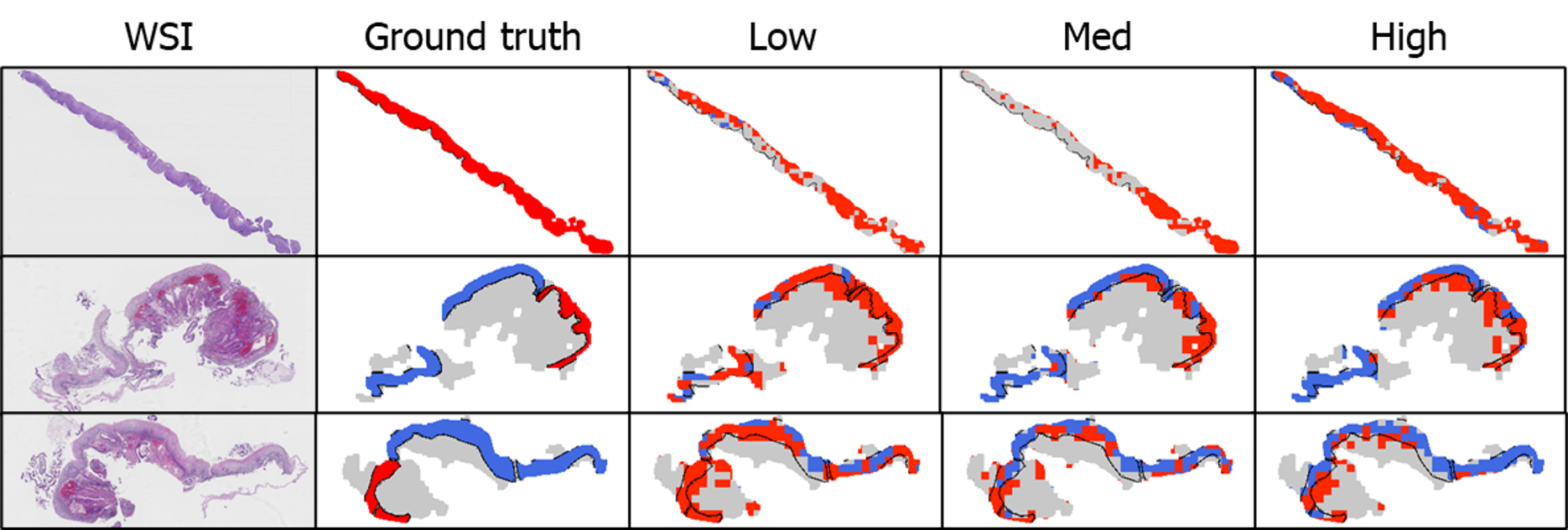}
\caption{Examples of segmentation results of the model trained using a WSI with the lowest, medium, and highest cluster entropy.
Gray, blue, and red indicate Non-Neop., LSIL, and HSIL, respectively.}
\label{fig:segmented_img}
\end{figure}

Fig. \ref{fig:segmented_img} shows examples of segmentation results for each condition. Our method (High) improved the segmentation. For example, in the first row, a wrongly estimated region (Non-Neop.) by medium entropy was correctly estimated as HSIL by our method.

\section{Conclusion}
In this paper, we proposed a cluster entropy metric for selecting an effective WSI from a target domain, which is used for semi-supervised domain adaptation.
Cluster entropy can measure how the image features of a WSI cover the distribution of all target data. Retraining a network using source data and a WSI with a high cluster entropy in the target domain improves the segmentation performance in the target domain. 
Experimental results demonstrated that our method achieved excellent performance for semi-supervised learning.
\vfill
\pagebreak
\noindent
{\bf Compliance with Ethical Standards: }
This study was performed in line with the principles of the Declaration of Helsinki. Approval was granted by the Ethics Committee of Tokyo University and Kyoto University.

\noindent
{\bf Acknowledgments: }
This work was supported by JSPS KAKENHI Grant Number JP21K19829.

\bibliographystyle{IEEEbib}
\bibliography{ISBI_latex}

\end{document}